\documentclass[times, review, 10pt]{elsarticle}

\usepackage{amssymb}
\usepackage{amsmath}
\usepackage{soul}
\usepackage{bbding}
\usepackage{xcolor}
\usepackage{multirow}
\usepackage{amssymb,amsbsy,subfigure,amsmath,amsfonts,multirow,color,url,bm,cases,eqparbox}
\usepackage{algorithm}
\usepackage{algpseudocode}
\usepackage{float}

\journal{Pattern Recognition}

\usepackage{cleveref}
\Crefname{figure}{Fig.}{Figures}

\usepackage{pifont}
\begin{document}
	
	\begin{frontmatter}
		
		
		
		\title{SpectralKAN: Weighted Activation Distribution Kolmogorov-Arnold Network for Hyperspectral Image Change Detection}
		
		
		\author{Yanheng Wang$^{a}$, Xiaohan Yu$^{b}$, Yongsheng Gao$^{c,*}$, Jianjun Sha$^{d}$, Jian Wang$^{e}$, Shiyong Yan$^{a}$, Kai Qin$^{a}$, Yonggang Zhang$^{d}$ and Lianru Gao$^{f}$}
		
		\cortext[cor1]{Corresponding author. Y. Gao. (e-mail: yongsheng.gao@griffith.edu.au)}  
		
		\address[inst1]{School of Environment and Spatial Informatics, China University of Mining and Technology, Xuzhou 221116, China}
		\address[inst2]{School of Computing, Macquarie University, Sydney, NSW 2109, Australia}
		\address[inst3]{School of Engineering and Built Environment, Griffith University, Nathan, QLD 4111, Australia}
		\address[inst4]{College of Intelligent Systems Science and Engineering, Harbin Engineering University, Harbin 150001, China}
		\address[inst5]{School of Intelligence Policing, China People's Police University, Langfang, 065000, China}
		\address[inst6]{Key Laboratory of Computational Optical Imaging Technology, Aerospace Information Research Institute, Chinese Academy of Sciences, Beijing 100094, China}

		
		\begin{abstract}
		Kolmogorov-Arnold networks (KANs) represent data features by learning the activation functions and demonstrate superior accuracy with fewer parameters, FLOPs, GPU memory usage (Memory), shorter training time (TraT), and testing time (TesT) when handling low-dimensional data. However, when applied to high-dimensional data, which contains significant redundant information, the current activation mechanism of KANs leads to unnecessary computations, thereby reducing computational efficiency. KANs require reshaping high-dimensional data into a one-dimensional tensor as input, which inevitably results in the loss of dimensional information. To address these limitations, we propose weighted activation distribution KANs (WKANs), which reduce the frequency of activations per node and distribute node information into different output nodes through weights to avoid extracting redundant information. Furthermore, we introduce a multilevel tensor splitting framework (MTSF), which decomposes high-dimensional data to extract features from each dimension independently and leverages tensor-parallel computation to significantly improve the computational efficiency of WKANs on high-dimensional data. In this paper, we design SpectralKAN for hyperspectral image change detection using the proposed MTSF. SpectralKAN demonstrates outstanding performance across five datasets, achieving an overall accuracy (OA) of 0.9801 and a Kappa coefficient ($\mathcal{K}$) of 0.9514 on the Farmland dataset, with only 8 \textit{k} parameters, 0.07 \textit{M} FLOPs, 911 \textit{MB} Memory, 13.26 \textit{s} TraT, and 2.52 \textit{s} TesT, underscoring its superior accuracy–efficiency trade-off. The source code is publicly available at https://github.com/yanhengwang-heu/SpectralKAN.
			
		\end{abstract}

		
	\end{frontmatter}

\section{Introduction}

Kolmogorov–Arnold Networks (KANs), neural architectures based on the Kolmogorov–Arnold representation theorem, have been proposed as alternatives to Multi-Layer Perceptrons (MLPs)~\cite{Hornik1989, Liu2024a}. Structurally, KANs are similar to MLPs, but they differ in their use of activation functions. While MLPs rely on a single activation function across all edges, KANs employ different, learnable activation functions on various edges, endowing them with stronger nonlinear representation capabilities. Although a single KAN layer with the same number of nodes contains significantly more parameters than an MLP layer, KANs require fewer layers to achieve superior feature extraction for low-dimensional data. This leads to a lower overall number of parameters (NP), fewer floating-point operations (FLOPs), reduced GPU memory usage (Memory), shorter training time (TraT) and testing time (TesT). However, KANs fail to perform well in handling high-dimensional data, such as hyperspectral image change detection.

\begin{figure}[!h]
	\centering
	\includegraphics[width=0.7\linewidth]{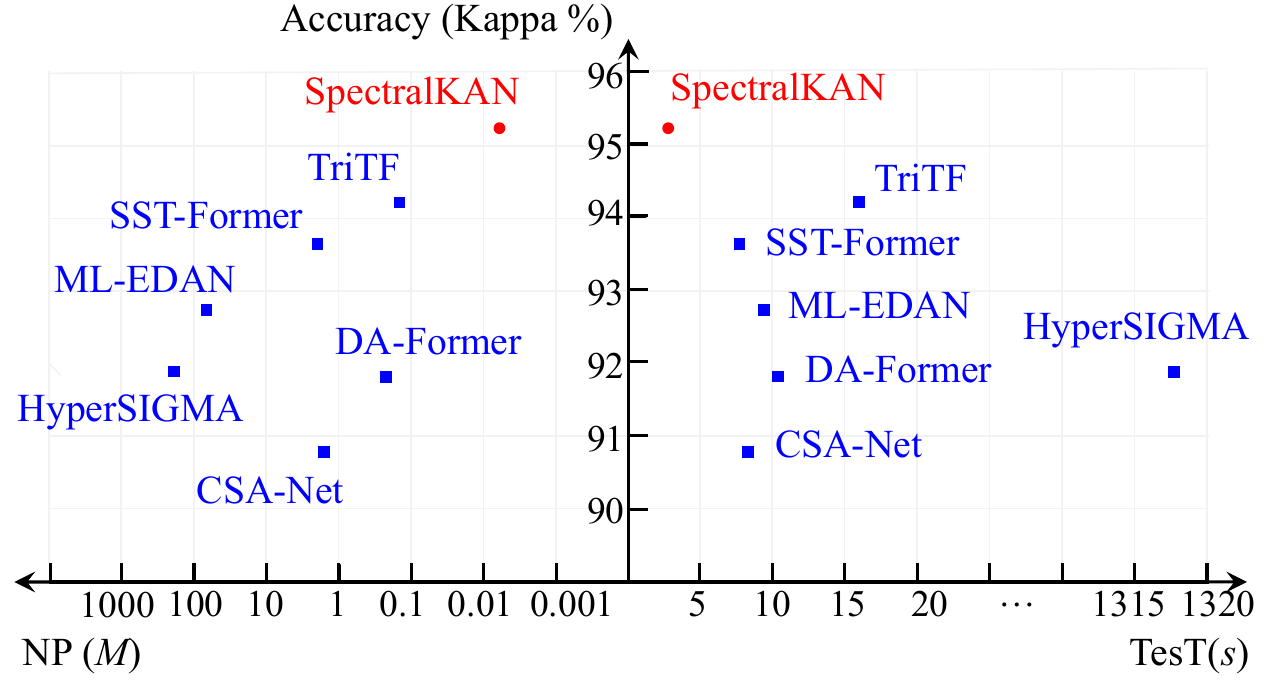}
	\caption{Performance comparison of state-of-the-art methods and the proposed SpectralKAN on the commonly used Farmland dataset.}
	\label{performance}
\end{figure}

Hyperspectral images are three-dimensional data that capture tens to hundreds of spectral bands from microwave to mid-infrared wavelengths~\cite{Wu2025, Jiang2025}. Hyperspectral image change detection involves analyzing multi-temporal images to obtain information about land changes~\cite{Qu2024a, Wang2021a}. Convolutional neural networks (CNNs), graph neural networks (GNNs), and transformers built on MLPs perform well on high-dimensional data~\cite{Li2017, Huang2021, Sellami2022}. However, their high accuracy comes with the cost of a large NP, high FLOPs, extensive Memory, and long TraT and TesT. These factors make computational efficiency crucial for high-dimensional data research, as they can significantly impact the feasibility and scalability of analysis. Given these challenges, advancing KANs technology to handle high-dimensional data is a promising approach.

The main challenges of advancing KANs for analyzing high-dimensional data are summarized as follows. 
\begin{itemize}
	\item A common approach to improving KANs is to replace MLPs in existing networks, such as U-Net, with KANs. However, KANs typically have more parameters than MLPs for the same number of nodes, thus negating their advantages in terms of computational efficiency and providing only limited accuracy improvements.
	\item KANs utilize a mechanism that involves multiple activations of one input node, leading to a substantial increase in NP and FLOPs for high-dimensional data. Additionally, high-dimensional data often contains significant information redundancy, resulting in many parameters being devoted to managing this redundant information.
	\item KANs are designed to accept one-dimensional inputs, necessitating the reshaping of high-dimensional data into a one-dimensional format. This reshaping process often leads to the loss of critical structural information inherent in the original data.
\end{itemize}

\begin{figure*}[!t]
	\centering
	\includegraphics[width=1\linewidth]{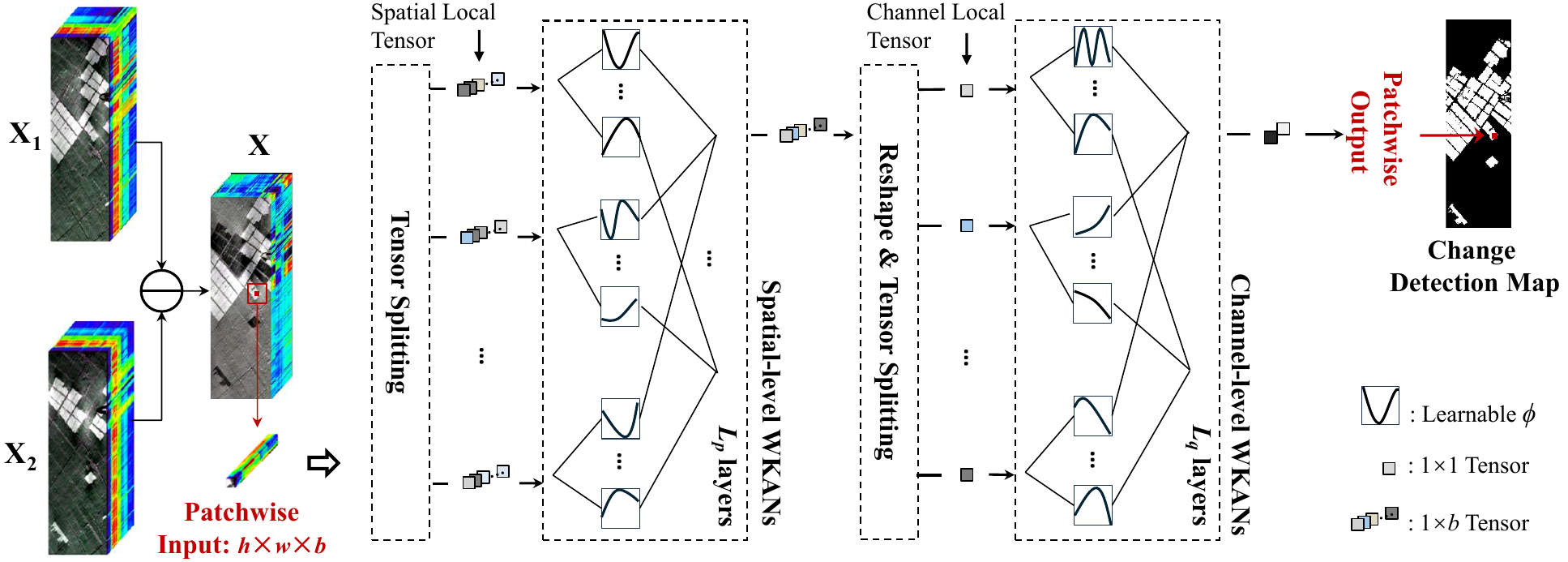}
	\caption{Flowchart of the SpectralKAN. SpectralKAN first splits each patch into multiple spatial local tensors $\{v_i\}_{i=1}^{h\times w}$. Global spatial features \textit{f} are then extracted using spatial-level WKANs within the MTSF. We further split \textit{f} into channel local tensors $\{f_e\}_{e=1}^{b}$. Channel-level WKANs in MTSF subsequently extract spectral features and classify them as either changed or unchanged.}
	\label{spectralkan}
\end{figure*}

In this paper, we take a significant step forward in advancing KANs for processing high-dimensional data. The weighted activation distribution Kolmogorov-Arnold networks (WKANs) are proposed, reducing the NP and FLOPs by reducing the number of activations per node. By employing a weighted activation distribution, we effectively capture the dependencies between input and output nodes, which in turn reduces the extraction of redundant information. We introduce a multilevel tensor splitting framework (MTSF) to extract features from each dimension of high-dimensional data and perform classification in the final layer. At each level, MTSF separates tensors and allocates them to different computation nodes, where WKANs are used to calculate relationships between local tensors, ultimately obtaining global features for that dimensional level. This tensor-parallel computation method not only improves computational efficiency but also accelerates processing times.

We demonstrate the effectiveness of our approach using hyperspectral image change detection as a case study. The proposed SpectralKAN, based on WKANs and MTSF, is validated on five datasets. As shown in Fig.~\ref{performance}, SpectralKAN outperformed state-of-the-art algorithms on the widely used Farmland dataset in terms of accuracy (Kappa), NP and TesT. SpectralKAN achieved the highest performance with the least NP and shortest TesT, highlighting its effectiveness and efficiency in high-dimensional data processing. There are three main contributions in this paper as summarized follows.
\begin{itemize}
	\item We introduce WKANs, an optimization of KANs for high-dimensional data, which reduce the number of activation functions per node, use weights to control their size, and distribute activation values to different output nodes. They compensate for information loss from fewer basic activations by extracting redundant information, significantly lowering the NP and FLOPs.
	\item We develop an MTSF, which addresses the structural information loss inherent in KANs by separating tensors along different dimensions and extracting features from each dimension. Additionally, the parallel tensor computation of MTSF enhances overall computational efficiency.
	\item We propose a novel SpectralKAN, advancing the pure KANs for high-dimensional data processing. This method eliminates the need to replace MLPs in classical networks, achieving higher accuracy while reducing NP, FLOPs, Memory, TraT, and TesT.
\end{itemize}

\section{Related Work}

\subsection{Kolmogorov-Arnold Networks}

Kolmogorov-Arnold representation theorem states that a multivariate function can be represented as the superposition of continuous functions of a single variable with two parameters. Kolmogorov-Arnold representation theorem has been used to view the neural network as a multivariate continuous function~\cite{Sprecher2002, Leni2013}. The depth and width of these networks have always been 2 and 2\textit{n}+1, respectively. They did not consider using back propagation to update the network. Based on Kolmogorov-Arnold representation theorem, Liu et al. \cite{Liu2024a} designed deeper and more flexible KANs, which have been proven to possess a stronger function fitting capability than MLPs.

KANs have quickly gained attention, leading to numerous applications. KANs were utilized to extract time-series information and proved their effectiveness in sequence feature extraction~\cite{VacaRubio2024, Genet2024, Huang2024}. Cross-dataset human activity recognition has been achieved based on KANs~\cite{Liu2024}. Wav-KAN~\cite{Bozorgasl2024} is a model that uses continuous or discrete wavelet transforms to fit continuous multivariate functions to get a better training speed, performance and computational efficiency than MLPs. Jamali et al.~\cite{Jamali2024} introduced HybridKAN for hyperspectral image classification. HybridKAN replaces the MLPs in 3D CNN, 2D CNN, and 1D CNN with KANs, thereby increasing the NP and FLOPs compared with CNNs, while also relying on principal component analysis (PCA) for dimensionality reduction, which leads to considerable spectral information loss. DeepOKAN \cite{Abueidda2024} not only replaces the B-splines in KANs with Gaussian radial basis functions but also substitutes the traditional MLPs in DeepONet with KAN-based architectures for modeling continuous operator mappings in complex engineering and mechanics problems. These design choices result in a substantially larger number of parameters than standard KANs. A U-KAN that combines U-Net and KANs is proposed to segment medical images~\cite{Liu2024}. Xu et al. \cite{Xu2024a} combined GCNs and KANs for recommendation tasks and used dropout to enhance the representational capability. A KCN~\cite{Cheon2024} that combines CNNs and KANs has been used for satellite remote sensing image classification. It validates the effectiveness of KANs for remote sensing image processing by replacing the MLPs in different CNN-based backbones with KANs, and demonstrates that KANs show better convergence.

The pure KAN-based methods have demonstrated good efficiency on low-dimensional data. However, studies on their application to high-dimensional data remain limited. Moreover, replacing MLPs with KANs in existing networks such as U-Net tends to increase the total NP due to the higher complexity of a single KAN layer, while yielding only marginal accuracy improvements. These studies have not extended the advantages of KANs to high-dimensional data, where computational efficiency is of paramount importance.

\subsection{Deep Learning for Hyperspectral Image Change Detection}

Deep learning demonstrates strong representational capabilities for high-dimensional data, achieving significant success in hyperspectral image processing. It is widely used to extract spectral-spatial features from hyperspectral images. Spectral and spatial attention networks employ learnable attention mechanisms, enabling the effective suppression of irrelevant spectral bands and spatial information~\cite{Wang2021, Hu2024, Yu2024}. Transformers have been specifically employed to learn and process spectral sequence information, thereby enhancing the capability to capture intricate spectral patterns and dependencies~\cite{Wang2022, Han2023}. Temporal information is crucial and various methods have been developed to detect changes between different temporal hyperspectral images. The commonly used method is subtracting~\cite{Guo2021, Ou2022} or concatenating~\cite{Zhang2023} multi-temporal images before feature extraction. Features were fused at different layers to obtain the multi-scale change features~\cite{Luo2023, Qu2024, Ning2025}. Long short-term memory network (LSTM) was utilized to learn temporal change information~\cite{song2018change, Bai2021, Shi2022}. Foundation models have been introduced into change detection, leveraging large-scale pretraining to enhance generalization across diverse datasets and scenarios~\cite{Wang2025, Gao2025}. For urgent tasks and real-time processing, computational efficiency is equally critical. However, most existing methods primarily focus on improving accuracy, while their performance with respect to NP, FLOPs, memory usage, TraT, and TesT remains unsatisfactory.
\section{Proposed Method}

\subsection{SpectralKAN Overview}
The overall flowchart of SpectralKAN is illustrated in Fig.~\ref{spectralkan}. Let $\mathbf{X_1}\in \mathbb{R}^{H \times W \times b}$ and $\mathbf{X_2}\in\mathbb{R}^{H \times W \times b}$ denote a pair of co-registered bi-temporal hyperspectral images. $\mathbf{X} = \mathbf{X_1}-\mathbf{X_2}$ is divided into multiple patches $\{x_i\}_{i=1}^{H \times W}$ with a stride of 1, where $ x \in \mathbb{R}^{h \times w \times b}$. $h \times w$ denotes the patch size, $b$ represents the number of spectral bands in each patch. $x$ undergoes tensor splitting, resulting in spatial local tensors $\{v_i\}_{i=1}^{h\times w}$, $v_i \in \mathbb{R}^{1 \times b}$. These local tensors are processed in parallel by the first level of the MTSF, where they interact to extract global spatial features $f \in \mathbb{R}^{1 \times b}$. Next, the spatial dimension is removed by reshaping $f$ into $\mathbb{R}^{b \times 1}$. $f$ is decomposed into $b$ channel-local tensors $\{f_e\}_{e=1}^{b}$. These tensors are then processed by the second level of the MTSF, producing a $1\times2$ output that captures global spatial-spectral features. Each level of the MTSF is built from WKANs. The $1\times2$ tensor indicates the probabilities of change and no-change, with the higher probability determining the change detection result. During testing, the results from all patches are combined to generate the final change detection map.

\subsection{Weighted Activation Distribution KANs}

\begin{figure}[!t]
	\centering
	\includegraphics[width=1\linewidth]{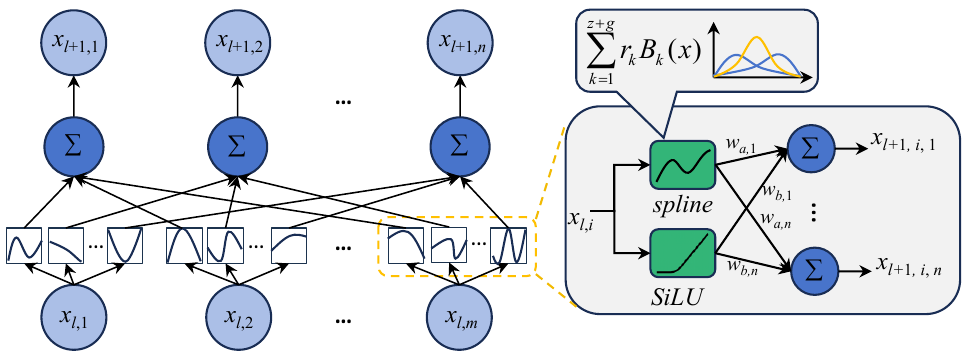}
	\caption{Structure of the \textit{l}-th WKAN layer with \textit{m} input nodes and \textit{n} output nodes.}
	\label{WKANs}
\end{figure}

WKANs are a special variant of KANs tailored for high-dimensional data, designed to extract features by learning activation functions. WKANs consist of $L$ layers $\mathbf{\Phi_{\textit{l}}}=\{\Phi_{\textit{l},1}, \Phi_{\textit{l},2}, \dots, \Phi_{\textit{l},\textit{m}}\}$, and the network as a whole can be represented as:
\begin{equation}
	\label{WaKAN}
	f(\mathbf{x})=\mathbf{x} \cdot \prod_{l=0}^{L-1} \bf \Phi_{\textit{l}}
\end{equation}
If the input to the $l$-th layer of WKANs is $\mathbf{x}_l \in \mathbb{R}^{m} $ and the output is $\mathbf{x}_{l+1} \in \mathbb{R}^{n}$, the computation within this layer is defined as:

\begin{equation}
	\label{WKAN_layer}
	\mathbf{x}_{l+1} = \begin{pmatrix}\Phi_{\textit{l},1} & \Phi_{\textit{l},2} & \dots &\Phi_{\textit{l},\textit{m}}\end{pmatrix}\begin{pmatrix} x_{\textit{l},1} \\ x_{\textit{l},2}\\ \vdots \\ x_{\textit{l},\textit{m}} \\ \end{pmatrix}
\end{equation}
In this equation, $\Phi_{\textit{l},\textit{i}}=\{\phi_{1}(\cdot),\phi_{2}(\cdot),\dots,\phi_{n}(\cdot)\}$ is the set of activation functions for $x_{\textit{l},\textit{i}}$, where $\textit{i} \in [0,\textit{m}]$. Each $\Phi_{\textit{l},\textit{i}}$ consists of two activation functions with weights such that 
\begin{equation}
	\label{Wkan}
	\begin{pmatrix}
		\phi_{1}(\cdot)\\
		\phi_{2}(\cdot)\\
		\vdots\\
		\phi_{n}(\cdot)\\
	\end{pmatrix} = \begin{pmatrix}
		w_{a,1} \\
		w_{a,2} \\
		\vdots\\
		w_{a,n} \\
	\end{pmatrix}\alpha(\cdot)+\begin{pmatrix}
		w_{b,1} \\
		w_{b,2} \\
		\vdots\\
		w_{b,n} \\
	\end{pmatrix} spline(\cdot)
\end{equation}
where $\alpha(\cdot)$ and $spline(\cdot)$ are basic activation functions. $\alpha(\cdot)$ refers to sigmoid linear unit (SiLU) function. $spline(\cdot)$  is composed of multiple B-spline basis functions:
\begin{equation}
	\label{spline}
	spline(x) = \sum_{i=1}^{z+g} r_i B_i(x)
\end{equation}
where $B_i(\cdot)$ is the \textit{i}-th B-spline basis function, $r_i$ is the corresponding weight, \textit{z} and \textit{g} are the degree and grid of the $spline(\cdot)$, respectively. The weights $w_a$ and $w_b$ scale the activation functions and distribute the activated input node information to different output nodes. For each input $x_{l,i}$ processed by $\Phi_{l,i}$, and \textit{n} outputs $\{x_{l+1,i,j}\}_{j=1}^n$ are produced. The detailed calculation process is illustrated in Fig.~\ref{WKANs}. Finally, the \textit{j}-th node of the output $x_{l+1, j}$ is computed by summing the contributions from all inputs:
\begin{equation}
	x_{l+1, j}=\sum_{i=1}^{m}x_{l+1,i,j}
\end{equation}

\subsection{Multilevel Tensor Splitting Framework}

Inspired by \cite{Wang2022,Wang2024,Chen2025}, we design the MTSF, which decomposes high-dimensional tensors into multiple lower-dimensional tensors and assigns each dimension-specific tensor to dedicated WKANs for feature extraction. Consider a \textit{s}-dimensional input $x \in \mathbb{R}^{d_1 \times d_2 \times \dots \times d_s}$ and an MTSF composed of $s$ WKANs layers. Initially, $x$ undergoes tensor splitting to obtain $d_1$ dimensional local tensors $\{v_i\}_{i=1}^{d_1}$, where $v_i \in \mathbb{R}^{d_2 \times \dots \times d_s}$. We define the first WKANs in the MTSF with $d_1$ input nodes and a single output node, processing all $d_1$ tensors in parallel, and aggregating their features into a unified representation:
\begin{equation}
	f = \prod_{l=0}^{L_1-1} \mathbf{\Phi}_{1,l} \begin{pmatrix}
		v_1\\
		v_2\\
		\vdots\\
		v_{d_1}\\
	\end{pmatrix}, f \in \mathbb{R}^{1 \times d_2 \times \dots \times d_s}
\end{equation}
where $\mathbf{\Phi}_{1,l}$ represents the activation functions of the $l$-th layer in the first WKANs, $L_1$ is the number of layers in the first WKANs within the MTSF. This structure ensures that $\{v_i\}_{i=1}^{d_1}$ are compressed into a single aggregated representation $f \in \mathbb{R}^{1 \times d_2 \times \dots \times d_s}$, capturing the global information along the first dimension. The first dimension is then removed by reshaping $f$ into $\mathbb{R}^{d_2 \times \dots \times d_s}$, facilitating the extraction of features along the second dimension. Tensor splitting is subsequently applied along the second dimension to obtain $d_2$ local tensors, which are then processed in parallel to extract features at this level. The second WKANs then process these tensors to capture global information along the second dimension, analogous to the first dimension. This process repeats for each subsequent dimension until global features for all dimensions are obtained. Finally, the last WKANs produces an output $y' \in \mathbb{R}^{1 \times c}$, where $c$ is the number of classes. The cross-entropy loss function is then used to calculate the difference between the predicted output $y'$ and the true label $y$, guiding the network’s optimization.

SpectralKAN is formulated as an MTSF with two WKANs for hyperspectral change detection, in which a hyperspectral image patch is decomposed into two sets of tensors corresponding to the spatial and spectral dimensions. The two WKANs, consisting of $L_p$ and $L_q$ layers, are employed to separately extract spatial and spectral representations. In the final layer, the output node count $c$ is set to two, representing the change and no-change classes. The pseudocode of SpectralKAN is shown in Algorithm~\ref{alg:mtsf_cd}.

\begin{algorithm}[H]
	\caption{SpectralKAN for Hyperspectral Image Change Detection}
	\label{alg:mtsf_cd}
	\begin{algorithmic}[1]
		\Require Bi-temporal hyperspectral images $\mathbf{X_1}, \mathbf{X_2} \in \mathbb{R}^{H \times W \times b}$
		\Ensure Change detection map $\mathbf{Y}$
		
		\State $\mathbf{X} \gets \mathbf{X_1} - \mathbf{X_2}$; initialize $\mathbf{Y} \gets \emptyset$
		\For{$i \gets 1 \ \textbf{to} \ H \times W$}
		\State $x_i \gets \text{Extract\_Patches}(\mathbf{X}, \text{patch\_size}=h \times w)$
		
		\State Split $x_i \to \{v_j\}_{j=1}^{h \times w}, \; v_j \in \mathbb{R}^{1 \times b}$
		
		\State $f \gets \text{Spatial-level WKANs}(\{v_j\})$
		\State Reshape $f$: $\mathbb{R}^{1 \times b} \to \mathbb{R}^{b \times 1}$
		\State Split $f \to \{f_e\}_{e=1}^{b}$
		
		\State $p_1, p_2 \gets \text{Channel-level WKANs}(\{f_e\})$
		\State $y \gets \arg\max(p_1, p_2)$
		\State $\mathbf{Y}_i \gets y$
		\EndFor
		\State \Return $\mathbf{Y}$
	\end{algorithmic}
\end{algorithm}

\subsection{Method Analysis}
\begin{figure}[!t]
	\centering
	\includegraphics[width=1\linewidth]{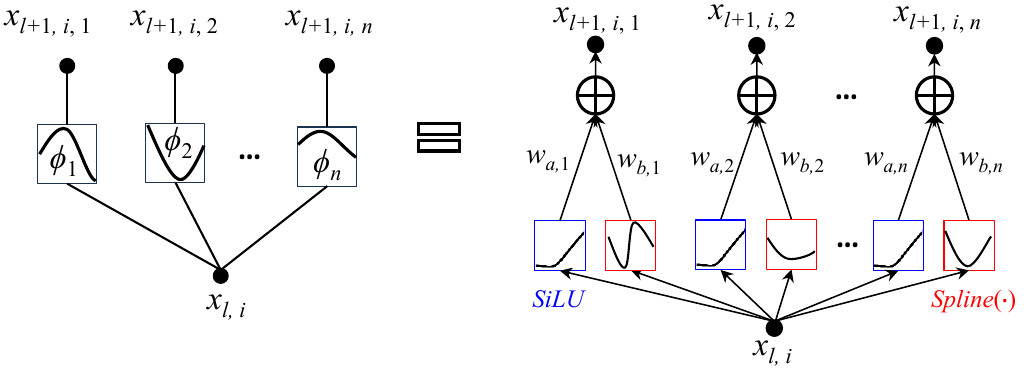}
	\caption{Structure of a single KAN layer with one input node and \textit{n} output nodes.}
	\label{KAN}
\end{figure}
\textbf{WKANs \textit{vs} KANs}: In KAN, each activation function $\phi(\cdot)$ is composed of a unique $\alpha(\cdot)$ and $spline(\cdot)$:

\begin{equation}
	\phi_{ij}(\cdot)=w_a\cdot \alpha_{ij}(\cdot)+w_b\cdot spline_{ij}(\cdot)
\end{equation}
where $i$ represents the $i$-th input node, and $j$ denotes the $j$-th output node. The detailed calculation process is illustrated in Fig.~\ref{KAN}. In comparison, WKAN uses a single $\alpha(\cdot)$ and a $spline(\cdot)$ for each input node across all output nodes, as indicated in Eq.~\ref{Wkan}. In both WKANs and KANs, the primary FLOPs in the activation function $\phi(\cdot)$ stem from computations within $\alpha(\cdot)$ and $spline(\cdot)$ and their multiplication with weights $w_a$ and $w_b$. Specifically, $\alpha(\cdot)$ incurs $O(4)$ FLOPs, while ${B_i(\cdot)}_{i=1}^{z+g}$ requires $O(4z(z+g))$ FLOPs. Each weight multiplication further requires $O(1)$ FLOPs. The main weights for $\phi(\cdot)$ include $w_a \in \mathbb{R}^{1}$, $w_b \in \mathbb{R}^{1}$, and $r_{i=1}^{z+g} \in \mathbb{R}^{z+g}$. For a layer with \textit{m} input nodes and \textit{n} output nodes, the FLOPs and NP are summarized as follows:
\begin{itemize}
	\item WKANs:\\
	- Flops: $O(2mn+4m(1+z(z+g)^2))$\\
	- NP: $O(2mn+m(z+g))$
	\item KANs:\\
	- Flops: $O(2mn+4mn(1+z(z+g)^2))$\\
	- NP: $O(2mn+mn(z+g))$
\end{itemize}

We can observe that a single WKAN layer has approximately \textit{n} times fewer NP and FLOPs compared to a single KAN layer. The activation mechanism in WKANs allows them to reduce the NP without compromising accuracy while still extracting additional features from redundant nodes.

\textbf{MTSF \textit{vs} WKANs}:
MTSF processes each dimension hierarchically and in parallel, reducing the number of nodes compared to WKANs. This leads to fewer activation functions at the edges and consequently reduces both FLOPs and NP. For example, SpectralKAN is an MTSF designed for hyperspectral data, processes input $x \in \mathbb{R}^{h \times w \times b}$ by spatial-level WKANs and channel-level WKANs with input node counts of $h \times w$ and $b$, respectively. In contrast, a single WKANs without tensor splitting would have $h \times w \times b$ input nodes. Assuming $L_p$ and $L_q$ are both set to 1, with output nodes being 1 in both SpectralKAN and a single WKAN layer, the computational aspects can be summarized as follows:
\begin{itemize}
	\item MTSF:\\
	- Flops: $O((hw+b)(2+4(1+z(z+g)^2))$\\
	- NP: $O((hw+b)(2+(z+g)))$
	\item WKANs:\\
	- Flops: $O(hwb(2+4(1+z(z+g)^2))$\\
	- NP: $O(hwb(2+(z+g)))$
\end{itemize}

The MTSF reduces the NP and FLOPs to approximately $(1/b + 1/hw)$ of those in WKANs. Moreover, MTSF enhances feature extraction by processing each dimension separately, leading to a better representation of high-dimensional data.

\section{Experiments}

\subsection{Datasets}
\footnotetext[1]{https://rslab.ut.ac.ir/data}
\footnotetext[2]{https://citius.usc.es/investigacion/datasets/hyperspectral-change-detectiondataset}

We conducted experiments using five publicly available datasets: Farmland\footnotemark[1], USA~\cite{Hasanlou2018}, River~\cite{Wang2018}, Bay Area\footnotemark[2], and Santa Barbara\footnotemark[2]. These datasets were captured by the Earth Observation-1 Hyperion hyperspectral sensor (EO-1) or the Airborne Visible/Infrared Imaging Spectrometer (AVIRIS). The changes in these datasets mainly involve land cover types and River variations. The satellite source (SS), imaging times (IT), cover land (CL), size (\textit{h}$\times$\textit{w}$\times$\textit{b}) and spatial resolution (SR) for five datasets are provided in Table~\ref{datasets}. For each dataset, we used 1\% of the pixels for training and the remaining pixels for testing, as detailed in Table~\ref{datasets_train_test}.

\begin{table*}[htb]
	\centering
	\caption{Details of Five Hyperspectral Image Change Detection Datasets\label{datasets}}
	\setlength{\tabcolsep}{3pt}
	\footnotesize
	\begin{tabular}{l c c c c c}
		\hline
		Dataset & SS & IT & CL & Size & SR \\
		\hline
		Farmland & EO-1 & 05. 2006 and 04. 2007 & Yancheng, China & 450$\times$140$\times$155 & 30\textit{m} \\
		River & EO-1 & 05. 2013 and 12. 2013 & Jiangsu, China & 431$\times$241$\times$198& 30\textit{m} \\
		USA & EO-1 & 05. 2004 and 05. 2007 & Hermiston, USA & 307$\times$241$\times$154& 30\textit{m} \\
		Bay Area & AVIRIS & 2013 and 2015 & Bay Area, USA & 600$\times$500$\times$224& 20\textit{m} \\
		Santa Barbara & AVIRIS & 2013 and 2014 & Santa Barbara, USA & 984$\times$740$\times$224 & 20\textit{m} \\
		\hline
	\end{tabular}
	
\end{table*}

\begin{table*}[htbp]
	\centering
	\caption{The Number of Training and Test Sets in the Five Datasets.\label{datasets_train_test}}
	\setlength{\tabcolsep}{3pt}
	\footnotesize
	\begin{tabular}{l c c c c c c c}
		\hline
		\multirow{2}{*}{Dataset} & \multirow{2}{*}{Unchanged} & \multirow{2}{*}{Changed} & \multirow{2}{*}{Unknown} & \multicolumn{2} {c} {Training Set} &\multicolumn{2} {c} {Testing Set} \\
		~&~&~&~&Unchanged & Changed & Unchanged & Changed\\
		\hline
		Farmland & 44723 & 18277 & 0 & 447 & 182 & 44276 & 18095 \\
		River & 101885 & 9698 & 0 & 1018 & 96 & 100867 & 9602 \\
		USA & 57311 & 16676 & 0 & 573 & 166 & 56738 & 16510 \\
		Bay Area & 34211 & 39270 & 226519 & 342 & 392 & 33869 & 38878 \\
		Santa Barbara & 80418 & 52134 & 595608 & 804 & 521 & 79614 & 51613 \\
		\hline
	\end{tabular}
\end{table*}

\begin{table*}[htb]
	\centering
	\caption{Comparison of OA, $\mathcal{K}$, NP, FLOPs, Memory, TraT and TesT with State-of-the-art Methods on Farmland. The Best Results are Highlighted In Bold.  \label{farmland}}
	\setlength{\tabcolsep}{3pt}
	\footnotesize
	\begin{tabular}{l c c c c c c c}
		\hline
		& OA & $\mathcal{K}$ & NP($k$) & FLOPs($M$) & Memory($MB$) & TraT($s$) & TesT($s$)\\
		\hline
		ML-EDAN & 0.97 & 0.9270 & 88130 & 275 & 3617 & 93.42 & 9.02\\
		SST-Former & 0.9743 & 0.9379 & 2498 & 145 & 1370 & 56.87 & 7.52\\
		CSANet & 0.9619 & 0.9075 & 2428 & 140 & 1561 & 63.9 & 7.85\\
		TriTF & 0.9754 & 0.9403 & 172 & 21 & 2406 & 49.02 & 16.06\\
		DA-Former & 0.9657 & 0.9174 & 398 & 30 & 1586 & 1444.43 & 10.46\\
		HyperSIGMA & 0.9661 & 0.9182 & 174596 & 29284 & 13585 & 868.32 & 1317.75\\
		SpectralKAN &\textbf{0.9801} & \textbf{0.9514} & \textbf{8} & \textbf{0.07} & \textbf{911} & \textbf{13.26} & \textbf{2.52} \\
		\hline
	\end{tabular}
\end{table*}

\begin{table*}[!h]
	\setlength{\tabcolsep}{3pt}
	\footnotesize
	\caption{Comparison of OA, $\mathcal{K}$, NP, FLOPs, Memory, TraT and TesT with State-of-the-art Methods on River. The Best Results are Highlighted In Bold. \label{river}}
	\centering
	\begin{tabular}{l c c c c c c c}
		\hline
		& OA & $\mathcal{K}$ & NP($k$) & FLOPs($M$) & Memory($MB$) & TraT($s$) &  TesT($s$)\\
		\hline
		ML-EDAN & 0.9484 & 0.6783 & 88526 & 285 & 3629 & 163.05 & 17.34\\
		SST-Former & 0.9644 & 0.7671 & 2520 & 148 & 1483 & 124.15 & 15.7\\
		CSANet & 0.9501 & 0.6762 & 2452 & 144 & 1586 & 29.9 & 10.84\\
		TriTF & 0.9699 & 0.8099 & 181 & 22 & 2486 & 95.45 & 31.42\\
		DA-Former & 0.9509 & 0.7041 & 409 & 32 & 1455 & 1477.47 & 20.11\\
		HyperSIGMA & 0.9622 & 0.7532 & 174629 & 29285 & 13471 & 1711.32 & 2374.74\\
		SpectralKAN & \textbf{0.9745} & \textbf{0.8366} & \textbf{9} & \textbf{0.09} & \textbf{961} & \textbf{25.55} & \textbf{5.1}\\
		\hline
	\end{tabular}
\end{table*}

\begin{table*}[!h]
	\centering
	\caption{Comparison of OA, $\mathcal{K}$, NP, FLOPs, Memory, TraT and TesT with State-of-the-art Methods on USA. The Best Results are Highlighted In Bold. \label{USA}}
	\setlength{\tabcolsep}{3pt}
	\footnotesize
	\begin{tabular}{l c c c c c c c}
		\hline
		& OA & $\mathcal{K}$ & NP($k$) & FLOPs($M$) & Memory($MB$) & TraT($s$) & TesT($s$)\\
		\hline
		ML-EDAN & 0.9400 & 0.8245 & 88121 & 275 & 3621 & 112.76 & 11.70\\
		SST-Former & 0.9431 & 0.8286 & 2498 & 145 & 1373 & 65.67 & 8.82\\
		CSANet & 0.9374 & 0.8167 & 2427 & 140 & 1561 & 77.5 & 9.73\\
		TriTF & 0.9563 & 0.8701 & 171 & 21 & 2418 & 59.69 & 19.89\\
		DA-Former & 0.9348 & 0.8169 & 398 & 30 & 1970 & 1269.04 & 12.41\\
		HyperSIGMA & 0.9454 & 0.8403 & 174595 & 29284 & 13583 & 989.03 & 1496.31\\
		SpectralKAN & \textbf{0.9591} & \textbf{0.8804} & \textbf{8} & \textbf{0.07} & \textbf{911} & \textbf{15.8} & \textbf{2.89}\\
		\hline
	\end{tabular}
\end{table*}

\subsection{Experimental Setup}

The SpectralKAN was implemented using PyTorch 2.3.0 with CUDA 11.8, and was trained on an Intel® Core™ i9-10900K CPU paired with 128 GB of RAM and NVIDIA TITAN RTX GPU. The operating system used was Ubuntu 20.04.1 LTS. In $spline(\cdot)$, $z$ and $g$ were set to 3 and 5, respectively, following commonly used values in existing KANs models. The layers $L_p$ and $L_q$ were set to 3 and 2. Each image patch was sized $5\times5$. The spatial-level WKANs in SpectralKAN comprised three layers with 25, 16, and 1 node(s) respectively, while the channel-level WKANs consisted of two layers with $b$ and 2 nodes. The training process involved 200 epochs with a batch size of 64. The Adam optimizer was used with a learning rate of 0.001, decayed by a factor of 0.9 every 10 epochs. The parameters ($w_a$, $w_b$, and $\textbf{r}$) were initialized using the Kaiming initialization method.

Six commonly used state-of-the-art methods, ML-EDAN~\cite{Qu2021a}, SST-Former~\cite{Wang2022}, CSANet~\cite{Song2022}, TriTF\cite{Wang2023}, DA-Former~\cite{Wang2023a}, and HyperSIGMA~\cite{Wang2025} were used as comparison methods. Performance was assessed using overall accuracy (OA) and Kappa ($\mathcal{K}$). The confusion matrix was employed to determine true positive (TP), true negative (TN), false positive (FP), and false negative (FN). OA represents the proportion of all pixels that are correctly classified:
\begin{equation}
	\label{OA}
	\rm OA = \dfrac{TP+TN}{TP+FP+TN+FN}
\end{equation}
$\mathcal{K}$ measures performance considering class imbalance and is given by:
\begin{equation}
	\label{kappa}
	\mathcal{K} = \dfrac{{\rm OA}-p_e}{1-p_e}
\end{equation}
\begin{equation}
	\label{pe}
	p_e = {\rm\dfrac{(TP+FP)(TP+FN)(TN+FN)(TN+FP)}{(TP+FP+TN+FN)^2}}
\end{equation}
Additionally, we compared the NP, FLOPs, Memory, TraT and TesT of different methods, as these metrics are crucial for practical applications in hyperspectral image change detection.

\begin{table*}[htb]
	\centering
	\caption{Comparison of OA, $\mathcal{K}$, NP, FLOPs, Memory, TraT and TesT with State-of-the-art Methods on Bay Area. The Best Results are Highlighted In Bold. \label{bayarea}}
	\setlength{\tabcolsep}{3pt}
	\footnotesize
	\begin{tabular}{l c c c c c c c}
		\hline
		& OA & $\mathcal{K}$ & NP($k$) & FLOPs($M$) & Memory($MB$) & TraT($s$) & TesT($s$)\\
		\hline
		ML-EDAN & 0.9634 & 0.9264 & 88766 & 291 & 3635 & 113.41 & 53.82\\
		SST-Former & 0.9661 & 0.932 & 2534 & 150 & 1581 & 78.89 & 52.71\\
		CSANet &\textbf{0.9826} & \textbf{0.9652} & 2468 & 147 & 1564 & 48.1 & 31.01\\
		TriTF & 0.9814 & 0.9626 & 186 & 22 & 2496 & 64.88 & 102.64\\
		DA-Former & 0.9807 & 0.9612 & 416 & 33 & 1991 & 1309.59 & 56.65\\
		HyperSIGMA & 0.9779 & 0.9555 & 174649 & 29286 & 13435 & 1033.53 & 1575.78\\
		SpectralKAN & 0.9641 & 0.9329 & \textbf{10} & \textbf{0.1} & \textbf{981} & \textbf{17.33} & \textbf{14.18}\\
		\hline
	\end{tabular}
\end{table*}

\begin{table*}[!h]
	\centering
	\caption{Comparison of OA, $\mathcal{K}$, NP, FLOPs, Memory, TraT and TesT with State-of-the-art Methods on Santa Barbara. The Best Results are Highlighted In Bold. \label{barbara}}
	\setlength{\tabcolsep}{3pt}
	\footnotesize
	\begin{tabular}{l c c c c c c c}
		\hline
		& OA & $\mathcal{K}$ & NP($k$) & FLOPs($M$) & Memory($MB$) & TraT($s$) & TesT($s$)\\
		\hline
		ML-EDAN & 0.9828 & 0.9636 & 88766 & 290 & 3631 & 210.73 & 128.59\\
		SST-Former & 0.9752 & 0.9478 & 2534 & 150 & 1544 & 149.85 & 134.62\\
		CSANet & 0.9916 & 0.9823 & 2468 & 147 & 1368 & 144.2 & 75.25\\
		TriTF & 0.9854 & 0.9693 & 186 & 22 & 2464 & 134.07 & 250.28\\
		DA-Former & \textbf{0.9920} & \textbf{0.9832} & 416 & 33 & 1984 & 1590.66 & 140.03\\
		HyperSIGMA & 0.9837 & 0.9658 & 174649 & 29286 & 13435 & 2068.93 & 2919.26\\
		SpectralKAN & 0.9776 & 0.9531 & \textbf{10} & \textbf{0.1} & \textbf{981} & \textbf{30.10} & \textbf{34.72}\\
		\hline
	\end{tabular}
\end{table*}

\subsection{Comparison with State-of-the-art Methods}

\begin{figure*}[!t]
	\centering
	\includegraphics[width=1\linewidth]{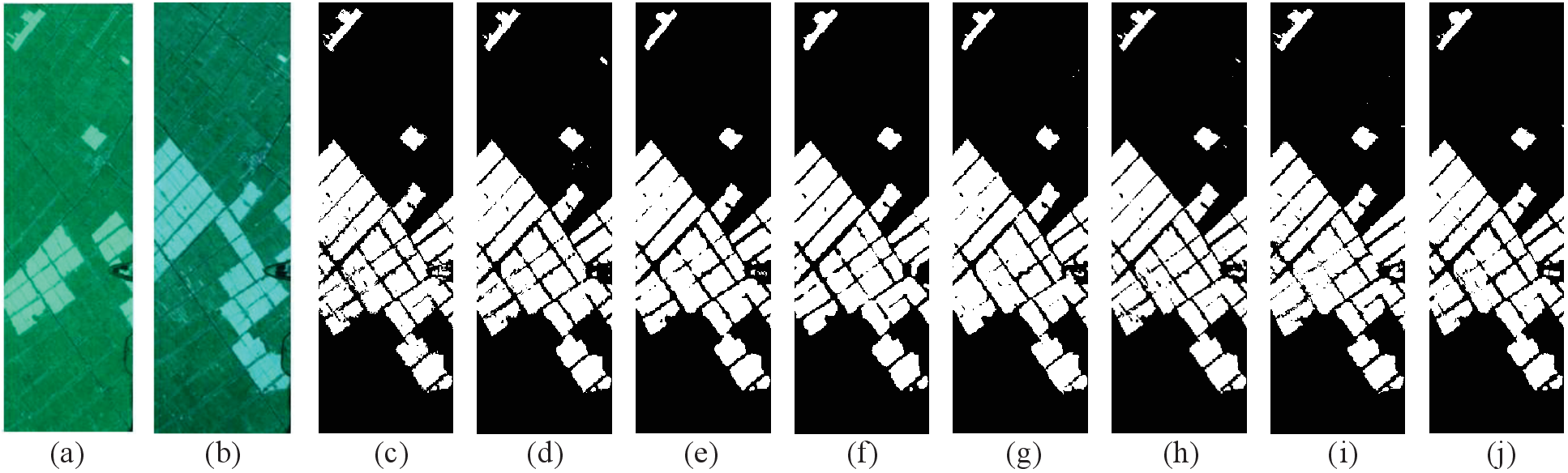}
	\caption{The results on Farmland datasets. (a) Before temporal hyperspectral images, (b) After temporal hyperspectral images, (c) Groundtruth, (d) ML-EDAN, (e) SST-Former, (f) CSANet, (g) DA-Former, (h) TriTF, (i) HyperSIGMA, (j) Ours.
		The white pixels are changed, and the black pixels are unchanged. }
	\label{farmland_result}
\end{figure*}

\begin{figure*}[htb]
	\centering
	\includegraphics[width=1\linewidth]{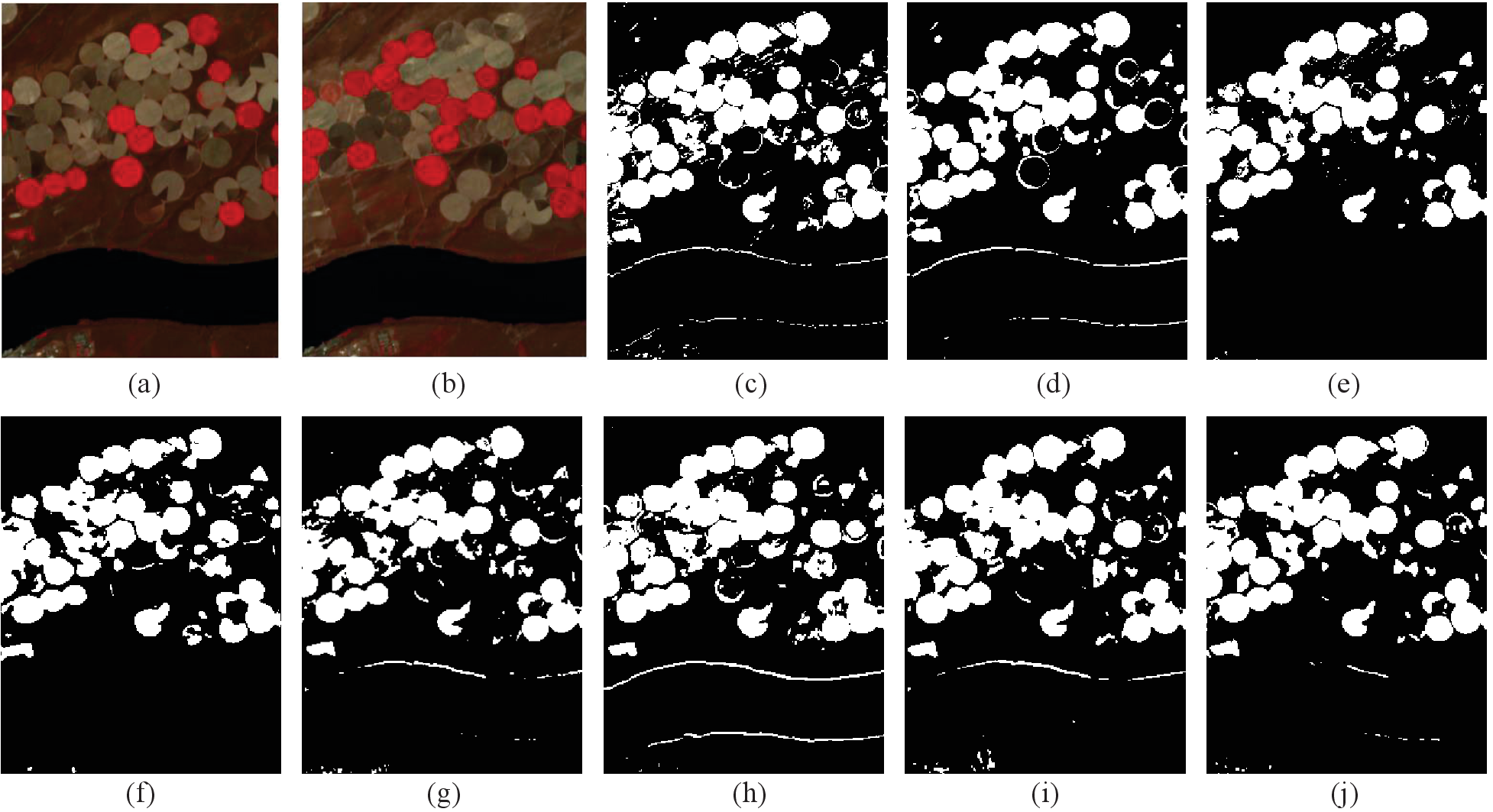}
	\caption{The results on USA datasets. (a) Before temporal hyperspectral images, (b) After temporal hyperspectral images, (c) Groundtruth, (d) ML-EDAN, (e) SST-Former, (f) CSANet, (g) DA-Former, (h) TriTF, (i) HyperSIGMA, (j) Ours. The white pixels are changed, and the black pixels are unchanged.}
	\label{USA_result}
\end{figure*}

\begin{figure*}[t]
	\centering
	\includegraphics[width=1\linewidth]{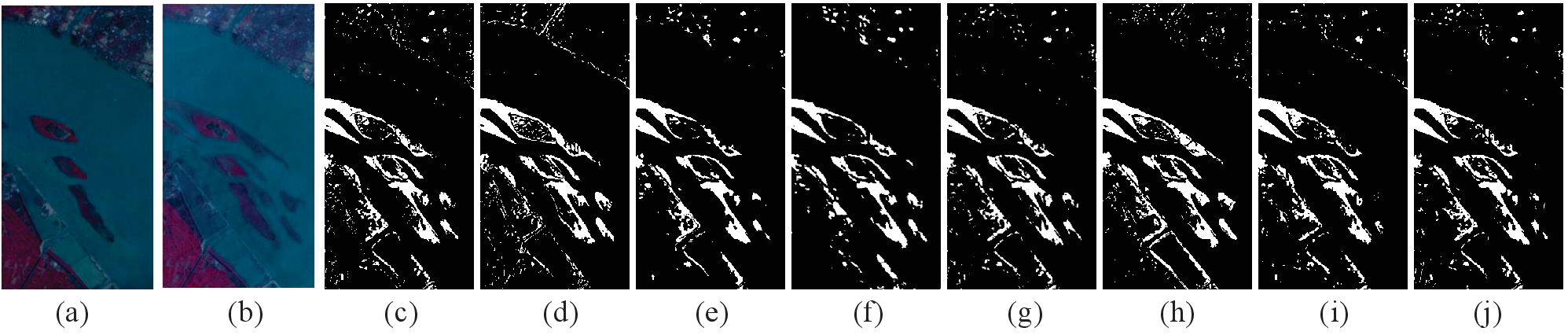}
	\caption{The results on River datasets. (a) Before temporal hyperspectral images, (b) After temporal hyperspectral images, (c) Groundtruth, (d) ML-EDAN, (e) SST-Former, (f) CSANet, (g) DA-Former, (h) TriTF, (i) HyperSIGMA, (j) Ours. The white pixels are changed, and the black pixels are unchanged.}
	\label{river_result}
\end{figure*}

\begin{figure*}[!h]
	\centering
	\includegraphics[width=1\linewidth]{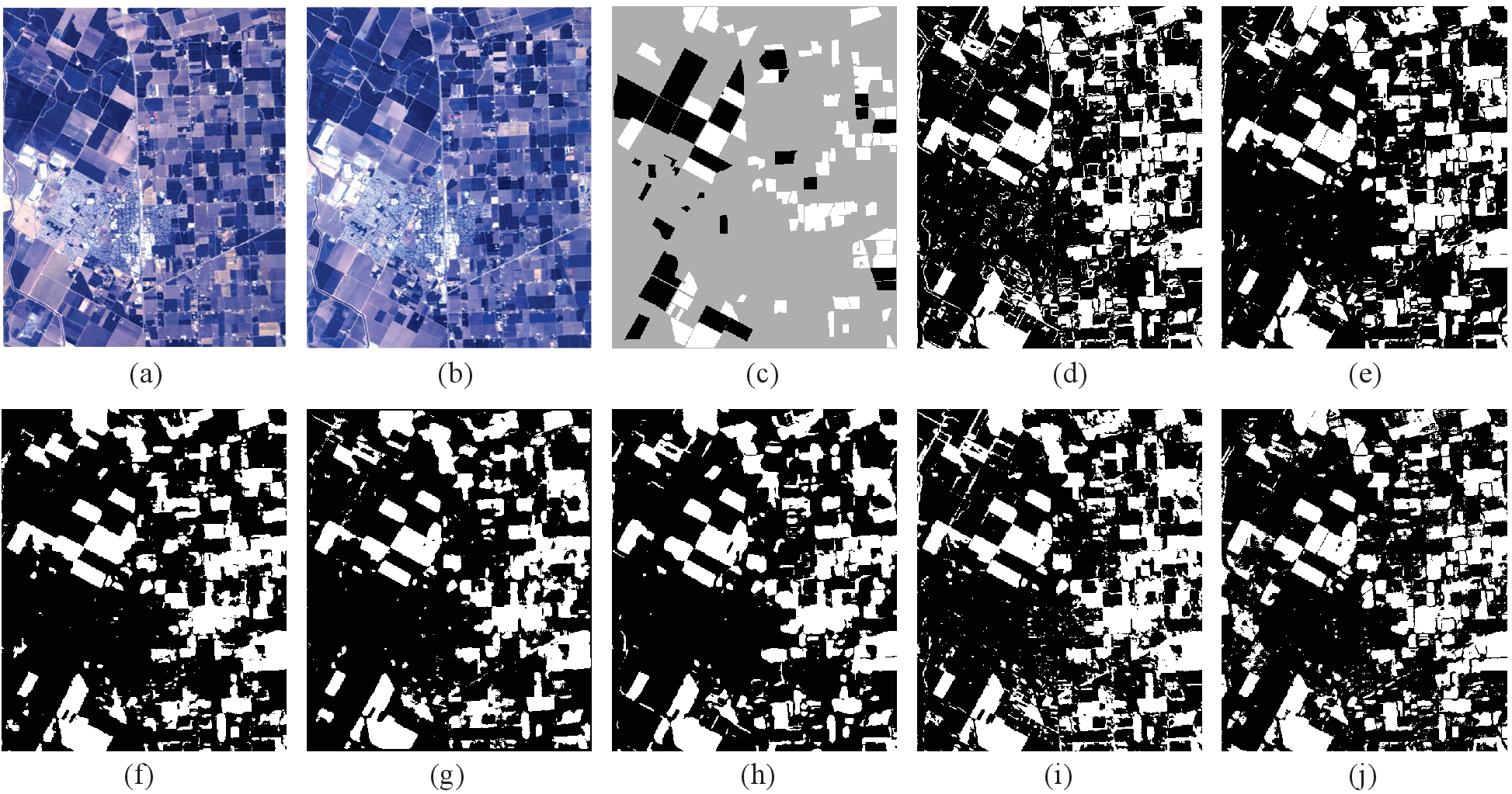}
	\caption{The results on Bay Area datasets. (a) Before temporal hyperspectral images, (b) After temporal hyperspectral images, (c) Groundtruth, (d) ML-EDAN, (e) SST-Former, (f) CSANet, (g) DA-Former, (h) TriTF, (i) HyperSIGMA, (j) Ours. The white pixels are changed, and the black pixels are unchanged. The gray pixels are unknown on ground truth.}
	\label{bayarea_result}
\end{figure*}

\begin{figure*}[htbp]
	\centering
	\includegraphics[width=1\linewidth]{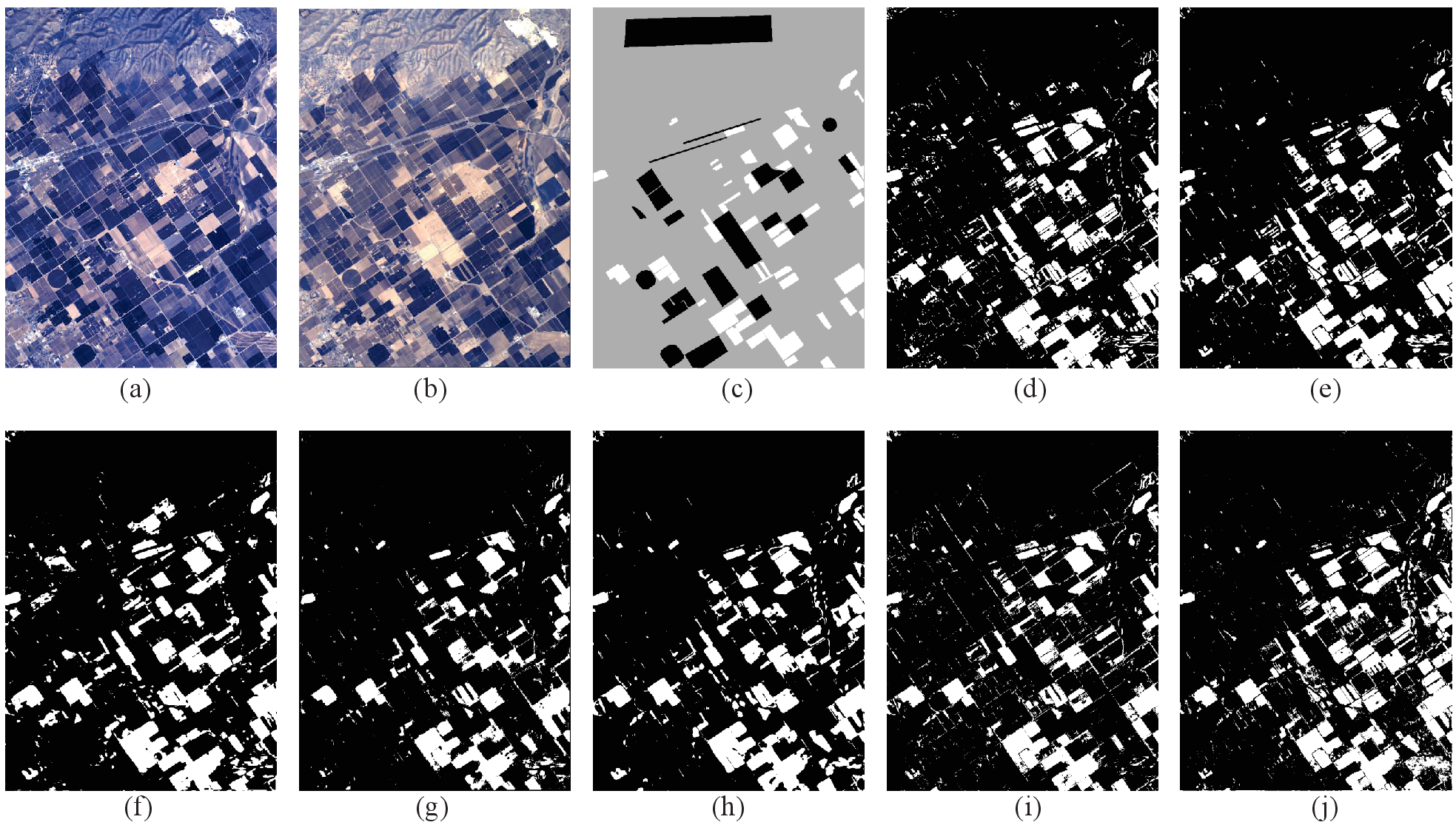}
	\caption{The results on Santa Barbara datasets. (a) Before temporal hyperspectral images, (b) After temporal hyperspectral images, (c) Groundtruth, (d) ML-EDAN, (e) SST-Former, (f) CSANet, (g) DA-Former, (h) TriTF, (i) HyperSIGMA, (j) Ours. The white pixels are changed, and the black pixels are unchanged. The gray pixels are unknown on ground truth.}
	\label{barbara_result}
\end{figure*}
Each experiment was repeated five times, and the average results are presented. Fig.~\ref{farmland_result}, \ref{river_result}, \ref{USA_result}, \ref{bayarea_result}, \ref{barbara_result} display the visual results for the Farmland, River, USA, Bay Area, and Santa Barbara datasets using state-of-the-art methods. The OA, $\mathcal{K}$, NP, FLOPs, Memory, TraT and TesT for these datasets are listed in Table~\ref{farmland},~\ref{river},~\ref{USA},~\ref{bayarea}, and~\ref{barbara}, respectively.

\textit{Results Analysis for the Farmland Dataset}: As illustrated in Fig.~\ref{farmland_result}, the ML-EDAN map shows considerable salt-and-pepper noise. SST-Former and DA-Former struggle with edge detection, while SST-Former and CSANet miss some alarm areas. TriTF and HyperSIGMA also exhibit minor salt-and-pepper noise. In contrast, SpectralKAN provides superior visual quality. Table~\ref{farmland} shows that SpectralKAN achieves the highest OA and $\mathcal{K}$, exceeding the second-best TriTF by 0.47\% and 1.11\%, respectively. Moreover, SpectralKAN demonstrates the lowest NP (8 \textit{k}), FLOPs (0.07 \textit{M}), memory usage (911 \textit{MB}), TraT (13.26 \textit{s}), and TesT (2.52 \textit{s}) among all methods, demonstrating its superior computational efficiency.

\textit{Results Analysis for the River Dataset}: Fig.~\ref{river_result} shows the visual results for the River dataset. ML-EDAN exhibited numerous false positives and false negatives. SST-Former, CSANet, and TriTF struggled with edge detection for small objects. DA-Former and HyperSIGMA missed several changes in small areas. In contrast, SpectralKAN showed improved visual results with fewer false negatives in small objects. In Table~\ref{river}, although ML-EDAN, SST-Former, CSANet, and DA-Former have shorter TesT, they show lower $\mathcal{K}$ and longer TraT. HyperSIGMA performs the worst across the last five evaluation metrics. SpectralKAN achieves the highest OA and $\mathcal{K}$, surpassing TriTF by 0.46\% in OA and 2.67\% in $\mathcal{K}$. SpectralKAN also demonstrates clear advantages, exhibiting the lowest NP (9 \textit{k}), FLOPs (0.09 \textit{M}), memory usage (961 \textit{MB}), TraT (25.55 \textit{s}), and TesT (5.1 \textit{s}) among all methods.

\textit{Results Analysis for the USA Dataset}: Fig.~\ref{USA_result} shows varying degrees of missed alarm areas across all methods. SST-Former and CSANet provide subpar visual results, while TriTF, DA-Former, and SpectralKAN perform well on large objects but struggle with narrow and small objects. Table~\ref{USA} confirms that SpectralKAN and TriTF are the top performers in OA and $\mathcal{K}$, while methods such as DA-Former exhibit the lowest accuracy on this dataset. SpectralKAN not only achieves the highest accuracy but also substantially reduces NP, FLOPs, Memory, TraT, and TesT compared to other state-of-the-art methods. In particular, SpectralKAN outperforms TriTF, with NP decreasing from 171 \textit{k} to 8 \textit{k}, FLOPs from 21 \textit{M} to 0.07 \textit{M}, Memory from 2418 \textit{MB} to 911 \textit{MB}, TraT from 59.69 \textit{s} to 15.8 \textit{s}, and TesT from 19.89 \textit{s} to 2.89 \textit{s}, demonstrating superior efficiency while maintaining the best performance.

\textit{Results Analysis for the Bay Area Dataset}: Fig.~\ref{bayarea_result} displays the visual results for the Bay Area dataset. While all methods effectively detected changes in labeled areas, only SST-Former and SpectralKAN provided detailed edge detection. Given that the annotated regions primarily consist of larger objects, the performance of SpectralKAN remains competitive, even though its accuracy is slightly lower compared to CSANet and other methods. Table~\ref{bayarea} indicates that CSANet, TriTF, and DA-Former achieve the top three OA and $\mathcal{K}$ scores. Although SpectralKAN lags behind by 3.23\%, 2.97\%, and 2.83\% in $\mathcal{K}$, respectively, it excels in detecting object edges and small objects, which are challenging for the other methods. Closer inspection reveals that most of SpectralKAN’s errors occur along change boundaries. This issue arises because spline functions are inherently smooth and continuous, making them better suited for modeling low-frequency and smoothly varying regions such as backgrounds or the interiors of large homogeneous objects. In contrast, boundaries in the Bay Area dataset contain substantial mixed pixels and high-frequency transitions. Consequently, spline-based activations may under-respond to these abrupt variations, leading to occasional missed detections along object edges. Even with this limitation, SpectralKAN produces clean boundary maps and delivers strong accuracy while keeping NP, FLOPs, memory, TraT, and TesT substantially lower than competing models.

\textit{Results Analysis for the Santa Barbara Dataset}: As shown in Fig.~\ref{barbara_result}, all methods perform well in the annotated regions. SpectralKAN demonstrates clear advantages in detecting small objects and delineating the boundaries of changed areas, underscoring its strong feature extraction capability. Table~\ref{barbara} further indicates that although SST-Former achieves slightly lower accuracy than some methods, its OA and $\mathcal{K}$ still reach 97.52\% and 94.78\%, respectively. By contrast, DA-Former, CSANet, and HyperSIGMA deliver higher change detection accuracy, but at the cost of substantially larger NP, Memory, and FLOPs, as well as longer TraT and TesT. SpectralKAN attains an OA of 97.76\% and a $\mathcal{K}$ of 95.31\%, which is about 3\% lower than the best-performing DA-Former in terms of $\mathcal{K}$, yet still represents competitive accuracy. Similar to the Bay Area dataset, the main source of error lies near object boundaries, where high-frequency mixed pixels cause the spline activation to undershoot sharp transitions, resulting in a small number of missed edge pixels. Despite this, SpectralKAN remains the most lightweight method, requiring only 10 \textit{k} parameters, 0.1 \textit{M} FLOPs, 981 \textit{MB} memory, and achieving the shortest TraT and TesT (30.10 \textit{s} and 34.72 \textit{s}) among all compared approaches.

In conclusion, ML-EDAN demonstrated relatively low OA and $\mathcal{K}$. SST-Former, DA-Former, and HyperSIGMA outperformed ML-EDAN, showing better results. CSANet excelled on the Bay Area and Santa Barbara datasets but underperformed on the other three. TriTF achieved second-highest accuracy across all datasets. All these methods exhibit higher NP, FLOPs, Memory, TraT and TesT. SpectralKAN achieved the best OA and $\mathcal{K}$ on the Farmland, River, and USA datasets, and also performed well on the Bay Area and Santa Barbara datasets. SpectralKAN is composed of only a few WKAN layers, resulting in a substantially reduced NP compared with the state-of-the-art methods. SpectralKAN integrates multiple activation functions, providing stronger nonlinear feature extraction capability. Consequently, the considerable reduction in NP does not compromise accuracy, and the lower NP directly translates into reduced FLOPs, Memory, TraT, and TesT, making SpectralKAN particularly suitable for deployment on devices with limited computational resources.

\begin{table*}[!h]
	\centering
	\caption{Ablation Experiment Results for WKANs and MTSF, Showing $\mathcal{K}$ and NP. The Best Results are Highlighted in Bold. \label{ablation}}
	\setlength{\tabcolsep}{1.5pt}
	\footnotesize
	\begin{tabular}{l c c c c c c c c c c c c}
		\hline
		&  \multirow{2}{*}{WKAN} &  \multirow{2}{*}{MTSF} &  \multicolumn{2}{c}{Farmland} & \multicolumn{2}{c}{River} & \multicolumn{2}{c}{USA} & \multicolumn{2}{c}{Bay Area} & \multicolumn{2}{c}{Barbara}\\
		&~&~&$\mathcal{K}$&NP($k$)&$\mathcal{K}$&NP($k$)&$\mathcal{K}$&NP($k$)&$\mathcal{K}$&NP($k$)&$\mathcal{K}$&NP($k$)\\
		\hline
		KANs&\XSolidBrush&\XSolidBrush& 0.9435 & 620 & 0.7638 & 792 & 0.8437 & 616 & 0.9109 & 896 & 0.9479 & 896\\
		WKANs&\Checkmark&\XSolidBrush& 0.9426 & 155 & 0.7648 & 198 & 0.8423 & 154 & 0.9185 & 224 & 0.9529 & 224\\
		MTSF-KANs&\XSolidBrush&\Checkmark& 0.9497 & 29 & 0.8219 & 36 & 0.878 & 29 & \textbf{0.9347} & 40 & \textbf{0.9565} & 40\\
		SpectralKAN&\Checkmark&\Checkmark& \textbf{0.9514} & \textbf{8} & \textbf{0.8366} & \textbf{9} & \textbf{0.8804} & \textbf{8} & 0.9329 & \textbf{10} & 0.9531 & \textbf{10}\\
		\hline
	\end{tabular}
\end{table*}

\subsection{Ablation Studies}

We conducted ablation studies to evaluate the effectiveness of the proposed WKANs and MTSF. Four model configurations were considered: the original KANs~\cite{Liu2024a}, the proposed WKANs, MTSF constructed with KANs (MTSF-KANs), and MTSF constructed with WKANs (SpectralKAN). The results across five datasets are summarized in Table~\ref{ablation}, using NP and $\mathcal{K}$ as evaluation metrics.

We observed that WKANs and SpectralKAN achieved approximately a fourfold reduction in NP compared with KANs and MTSF-KANs, respectively. Despite this substantial reduction, $\mathcal{K}$ remained largely stable across datasets, indicating that WKANs effectively suppress redundant representations in high-dimensional data without compromising information integrity. Furthermore, comparisons between KANs and MTSF-KANs, as well as between WKANs and SpectralKAN, showed that MTSF achieved more than a twentyfold reduction in NP while simultaneously improving $\mathcal{K}$. These results demonstrate that MTSF not only enhances high-dimensional feature extraction but also substantially reduces computational cost.

\subsection{Effect of Hyperparameters}

To study the effect of hyperparameters on SpectralKAN, we focused on the number of nodes and hidden layers, as well as the number of training samples. Other hyperparameters were set based on prior work and empirical experience.

\begin{figure}[htbp]
	\centering
	\includegraphics[width=1\linewidth]{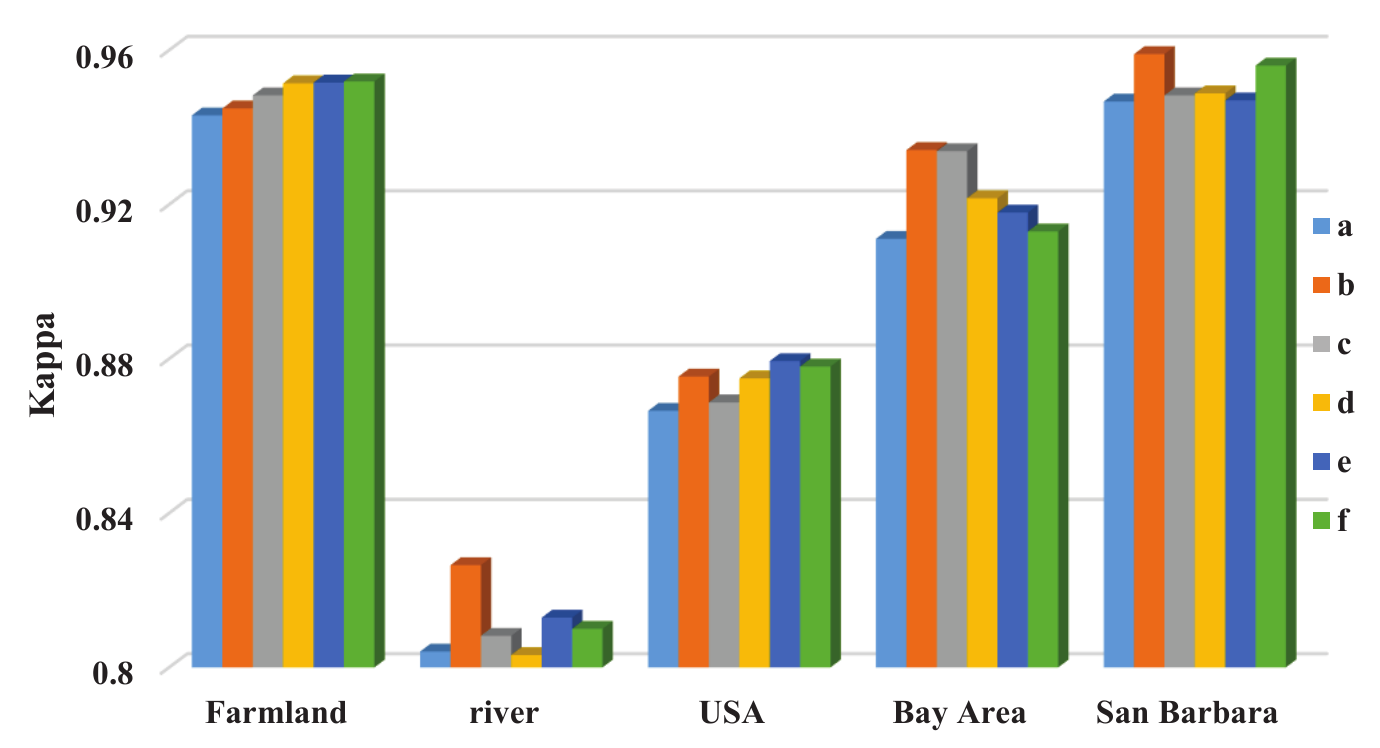}
	\caption{The $\mathcal{K}$ of different nodes and layers on five datasets.}
	\label{notes}
\end{figure}

\begin{figure}[htbp]
	\centering
	\includegraphics[width=1\linewidth]{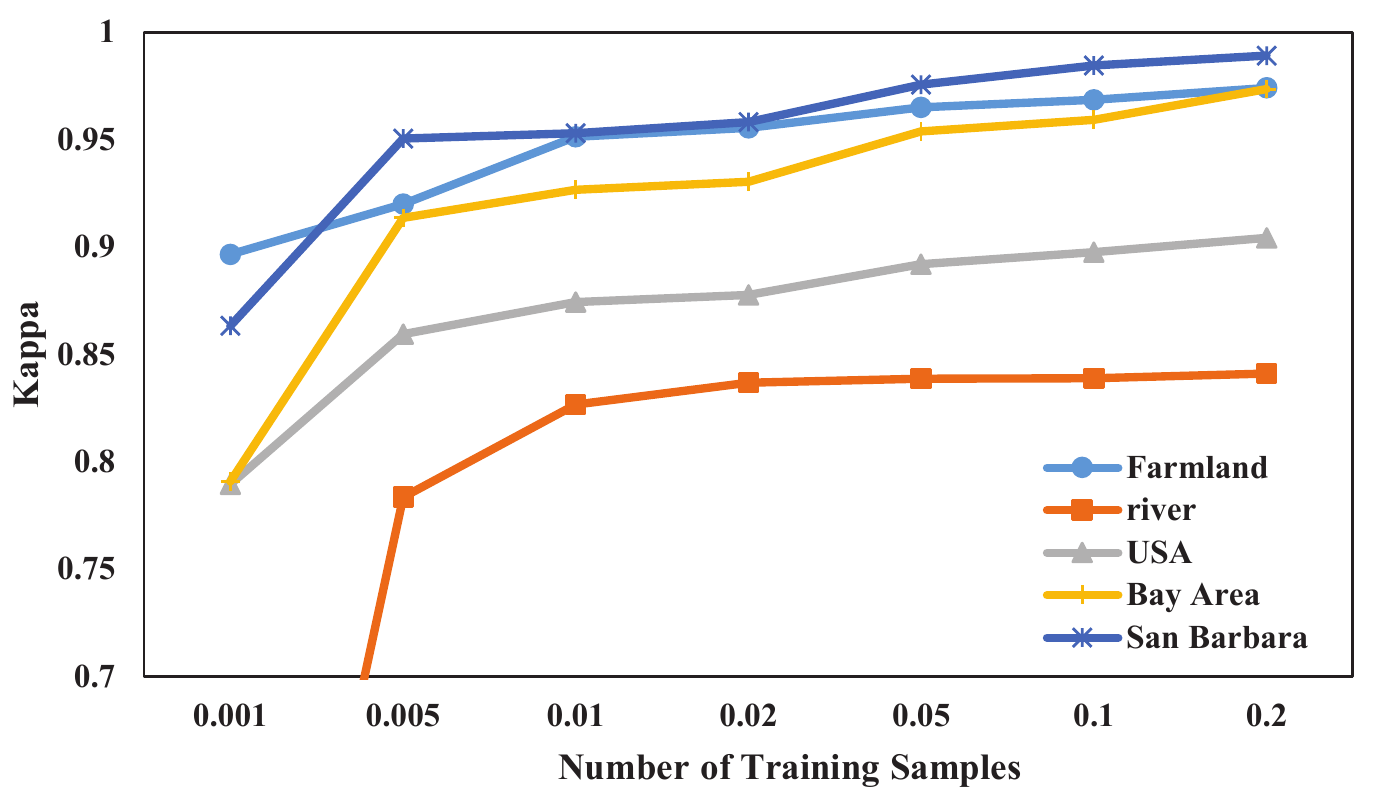}
	\caption{The $\mathcal{K}$ of different training set number on five datasets.}
	\label{trainingset}
\end{figure}

Different numbers of nodes and hidden layers in SpectralKAN result in different NP and accuracy. We selected six different combinations of nodes and layers to identify the optimal structure: a: [25,1], [\textit{b},2],
b: [25, 16, 1], [\textit{b}, 2],
c: [25, 1], [\textit{b}, 16, 2],
d: [25, 16, 1], [\textit{b}, 16, 2],
e: [25, 64, 1], [\textit{b}, 16, 2],
and f: [25, 16, 1], [\textit{b}, 64, 2]. In a–f, the first set specifies the number of nodes in each layer of the spatial-level WKANs (e.g., [25,1] indicates 25 nodes in the first layer and 1 node in the second layer), while the second set provides the corresponding layer-wise node configuration of the channel-level WKANs. From a to f, the number of nodes or layers gradually increases, accompanied by an increase in NP. The $\mathcal{K}$ of experiments can be seen in Fig.~\ref{notes}. In the Farmland dataset, as the number of nodes or layers increases, the $\mathcal{K}$ also gradually increases. In the other datasets, however, no consistent pattern is observed. Considering both NP and $\mathcal{K}$, we determine that b is the best configuration. 

Different tasks have varying requirements for number of training samples. {0.1\%, 0.5\%, 1\%, 2\%, 5\%, 10\%, 20\%} of the pixels as training sets to analyze their impact on SpectralKAN. The results are shown in Fig.~\ref{trainingset}. We observe that with a larger number of training samples, the accuracy of change detection increases. 

\section{Conclusion}
In this paper, we propose WKANs and MTSF to advance KANs for processing high-dimensional data. We apply these innovations to hyperspectral image change detection by developing the SpectralKAN. Extensive experiments on five datasets demonstrate that SpectralKAN achieves an average OA of 97.11\% and a $\mathcal{K}$ of 91.09\%. Meanwhile, it substantially reduces NP, FLOPs, and Memory, achieving TesT up to two times faster than the best baseline method. Overall, SpectralKAN combines competitive accuracy with high efficiency, making it suitable for scenarios with limited computational resources. But spline-based activations tend to produce overly smooth responses in the high-frequency regions typically found along hyperspectral object boundaries, which may lead to missed detections at change edges. In future work, we plan to address this limitation by incorporating frequency-domain transformations, such as the Discrete Cosine Transform, to enhance the model’s ability to capture high-frequency boundary information. Additionally, we will extend the MTSF to other high-dimensional tasks, such as 3D point clouds and video data, to further validate its effectiveness and broaden its applications.

\section{Declaration of Interest Statement}
The authors declare that they have no known competing financial interests or personal relationships that could have appeared to influence the work reported in this paper.

\section{Acknowledgement}
This work was supported in part by the National Key Laboratory on Electromagnetic Environmental Effects and Electro-optical Engineering under Grant KY3240020001 and the Aerospace Science and Technology Innovation Development Fund under Grant ZY0110020009.

\bibliographystyle{elsarticle-num}
\bibliography{reff1}


\end{document}